\documentclass[11pt,a4paper]{article}
\usepackage[nohyperref]{acl2019}
\usepackage{times}
\usepackage{latexsym}

\usepackage{url}
\usepackage{soul}
\usepackage{multirow}
\usepackage{float}
\usepackage{graphicx}
\usepackage{subcaption}

\aclfinalcopy 
\def\aclpaperid{68} 

\newcommand{\B}[1]{\textbf{#1}}
\newcommand{\clinquote}[1]{\texttt{#1}}

\title{Classifying the reported ability in clinical mobility descriptions}

\author{
    Denis Newman-Griffis$^{1,2}$\thanks{\ \ These authors contributed equally to this work.},
    Ayah Zirikly$^1$\footnotemark[1],
    Guy Divita$^1$\footnotemark[1],
    Bart Desmet$^1$\\
    $^1$Rehabilitation Medicine Department, Clinical Center, National Institutes of Health, Bethesda, MD\\
    $^2$Department of Computer Science and Engineering, The Ohio State University, Columbus, OH\\
    \texttt{ \{denis.griffis, ayah.zirikly, guy.divita, bart.desmet\}@nih.gov }
}

\date{}

\begin{document}
\maketitle
\begin{abstract}
    Assessing how individuals perform different activities is key information
for modeling health states of individuals and populations. Descriptions of
activity performance in clinical free text are complex, including syntactic
negation and similarities to textual entailment tasks. We explore a variety
of methods for the novel task of classifying four types of assertions about
activity performance: \textit{Able}, \textit{Unable}, \textit{Unclear}, and \textit{None} (no information). We
find that ensembling an SVM trained with lexical features and a CNN achieves 
77.9\% macro F1 score on our task, and yields nearly 80\% recall on the rare \textit{Unclear} and \textit{Unable} samples. Finally, we highlight several
challenges in classifying performance assertions, including
capturing information about sources of assistance, incorporating syntactic structure and negation scope, and handling new modalities at test time.
Our findings establish a strong baseline for this novel task, and identify intriguing areas for further research.
\end{abstract}
\section{Introduction}
\label{sec:intro}

Information on how individuals perform activities and participate in social
roles informs conceptualizations of quality of life, disability, and social
well-being. Importantly, activity performance and role participation are highly
dependent on the environment in which they occur; for example, one individual
may be able to walk around an office without issue, but experience severe
difficulty walking along mountain paths. Thus, determining what level of
performance an individual can achieve for activities in different environments
is critical for identifying ability to meet work requirements, and designing
public policy to support the participation of all people.

However, the interaction between individuals and environments makes modeling
performance information a complex task. Assessments of activity performance
within clinical healthcare settings are typically recorded in free text
\cite{Bogardus2004,Nicosia2019}, and exhibit high flexibility in structure.
Syntactic negation can be present, but is not necessarily indicative of
inability to perform an action; for example, \clinquote{Patient can walk with
rolling walker} and \clinquote{Patient cannot walk without rolling walker} are
both likely to be used to assert the ability of the patient to walk with the
use of an assistive device. Information about performance may also be given
without a clear assertion, as in \clinquote{the cane makes it difficult to walk}.
Thus, extraction of performance information must not only distinguish between
positive and negative assertions, but also those which cannot be clearly evaluated.

To the best of our knowledge, this is the first work to explore assertions of
activity performance in health data. We explore a variety of methods for classifying
assertion types, including rule-based approaches, statistical methods using
common text features, and convolutional neural networks. We find that machine
learning approaches set a strong baseline for discriminating between four assertion 
types, including rare negative assertions. While this work focuses on a
relatively constrained and homogeneous corpus, error analysis suggests several
broader directions for future research on classifying performance assertions.

\section{Related Work}
\label{sec:related}

Though this is the first work focusing on the polarity of activity performance, three areas of prior work are particularly relevant to this research.

The first is concerned with applying NLP techniques and linguistic annotation to information about whole-person function, particularly activity performance.
\citet{harris2003term} experimented  with  term  extraction for the purpose of terminology discovery to support information retrieval 
relating to functioning, disability and health, 
using linguistic, n-gram and hybrid techniques. 
\citet{bales2005extending} and \citet{kukafka2006human} modified and applied the MedLEE NLP Extraction tool to code Rehabilitation Discharge Summaries using ICF \cite{ICF} encodings. 
\citet{kuang2015representation} studied UMLS term coverage of functional status terms found in VA clinical notes and in social media sources, reporting that there is a need to extend existing terminologies to cover this area. 
Finally, \citet{Thieu2017} reported on an effort to build an annotated corpus of Physical Therapy (PT) notes from the Clinical Center of the National Institutes of Health (NIH) with functional status information. This corpus was also used for an investigation into using named entity recognition (NER) techniques to extract information about patient mobility \cite{newman2018embedding}.

The second area is research on negation. Negation detection is a well-researched area \cite{Morante2012}, and both negation and uncertainty have historically been studied in the clinical NLP context \cite{Mowery2012,Peng2018}. Previous work studied the use of incorporating dependency parsers to help in identifying the scope~\cite{sohn2012dependency, mehrabi2015deepen}. Recent work in this area involves the use of neural network models, where Long Short-Term Memory (LSTM), or variations of it, yielded competitive results on negation (cues and scope) detection~\cite{taylor2018role}. 

One highly-related work to ours is \citet{Wu2014}, which investigates detection of binary semantic negation status (i.e., the presence or absence of a finding, as opposed to syntactic negation) for clinical findings in EHR text. However, as Action Polarity is defined in terms of the interaction between an individual and a specific environment, it adds a layer of complexity to non-interactive physiological observations. \citet{gkotsis2016don} investigate using parsing-based scoping limitations for negation detection in complex clinical statements, though their focus is specifically on mentions of suicide.

Finally, classifying the assertion status of activity performance descriptions bears similarities to the problem of recognizing textual entailment (RTE) \cite{Dagan2006,Marelli2014}. RTE asks whether a given premise entails a specific hypothesis, and has historically been pursued in the general domain, though, recent efforts have developed datasets in biomedical literature \cite{BenAbacha2015,BenAbacha2016} and in clinical text \cite{Romanov2018}. Our task, by asking whether a given description entails ability to perform an action in the an environment, is more constrained than RTE, but poses a related research challenge.
\section{Data}

We use an extended version of the dataset initially described by
\citet{Thieu2017}, consisting of 400 English-language Physical Therapy initial
assessment and reassessment notes from the Rehabilitation Medicine Department
of the NIH Clinical Center. These text documents have been annotated to identify
descriptions and assessments of mobility status, typically including one or
more specific Actions; for example, \clinquote{Pt \ul{walked} 300' with rolling
walker} (Action underlined).

Each Action annotation was assigned one of four Polarity values, indicating
what (if any) information the containing mobility description provides about
the subject's ability to perform the given Action in the context of any described
environmental factors.\footnote{
    It is important to note that the Polarity label is dependent on the
    environmental factors described. For example, an individual may be able
    to walk a certain distance using an assistive device such as a rolling
    walker, but unable to walk that same distance independently.
} The Polarity labels are defined in the following paragraphs.

\textbf{Able} The subject is able to complete the activity in the
          environment described. For example, \clinquote{She states she can
          \ul{walk} 20 minutes before tiring}; in the case of \clinquote{now
          requires assistance of one person with \ul{transfers}}, it is unknown
          whether the patient can perform the action independently, but they
          are able to do so with the assistance described.
          
    \textbf{Unable} The subject is not able to complete the activity in
          the environment described; for example, \clinquote{He is unable to
          \ul{walk}}. More specific information may also be included, as in
          \clinquote{Pt is now unable to \ul{walk} more than 50 feet}.
          
    \textbf{Unclear} Some information is provided about the subject's
          ability to perform the action, but not enough to make a definitive
          positive or negative judgment. For example, in \clinquote{The cane
          makes it difficult to \ul{walk}}, it is undetermined whether the
          subject can or cannot walk. This label also includes some cases of
          negated environmental factors; for example, \clinquote{unable to
          \ul{propel} wheelchair independently}.
          
    \textbf{None} No direct information about ability to perform the
          action is provided. Common examples of this label refer to a scale
          that is either unavailable or distant in the document, as in
          \clinquote{\ul{Ambulation}: 1}. Other cases refer to a specific
          aspect of performing an action, without evaluation, as in
          \clinquote{tendency during \ul{gait} to quickly extend the leg from
          swing to stance}.

\begin{table}[t]
    \centering
    \begin{tabular}{r|cc|c}
        Label&Train&Test&Total\\
        \hline
        Able   &1,536&446&1,982\\
        Unable &   54& 23&   77\\
        Unclear&  158& 48&  206\\
        None&1,784&478&2,262\\
        \hline
        Total  &3,532&995&4,527\\
    \end{tabular}
    \caption{Number of samples with each Polarity label in train and test data.}
    \label{tbl:dataset}
\end{table}

We randomly split the 400 documents into 320 training records and 80 testing
records, stratified by distribution of Polarity labels.
Table~\ref{tbl:dataset} provides frequencies of each label in these splits.

\section{Methods}
\label{sec:approach}

We investigate a variety of methods to classify the Polarity values of Action annotations. Rule-based methods have been used to great effect in clinical information extraction \cite{Kang2013,chapman2007context}, and form an important baseline for our task. We also make use of several common machine learning methods, such as support vector machines and $k$-nearest neighbors, along with more recent neural models such as convolutional neural networks (CNN). Finally, we experiment with ensembled combinations of our best-performing models. These approaches are described in the following subsections.

\begin{table*}[t]
\centering
{\small
\begin{tabular}{lllp{7cm}}
Slot criteria&Value criteria&Assigned Polarity& Example                               \\
\hline
Asserted Action        & Asserted Evidence       & Able                              & \clinquote{Transfers: Independent}                         \\
Asserted Action        & Negated Evidence        & Unable                            & \clinquote{Transfers: Unable}                              \\
Negated Action         & Negated Evidence        & Able                              & \clinquote{Difficulty Walking: No}                         \\
Negated Action         & Asserted Evidence       & Unable                            & \clinquote{Unable to Walk: yes}                            \\
Asserted Action        & Numbers                 & Unclear                           & \clinquote{Transfers: 4}                                   \\
Asserted Action        & No context evidence     & Unclear                           & \clinquote{Sit to stand: minimal assist}                   \\
Asserted Action        & No value                & None                              & \texttt{Stand to sit:}    \\
Multiple Actions       & Doesn't matter          & None                              & \texttt{Difficulty with chores, shopping, driving: Yes} \\  
\end{tabular}
}
\caption{Table of slot:value rules for Action Polarity}
\label{tab:slotValueActionPolarityTable}
\end{table*}

\subsection{Rule-based}
\label{subsec:rule}
A UIMA \cite{ferrucci2004uima} based pipeline was constructed to identify action polarity from components of v3NLP-Framework \cite{divita2016v3nlp}.  
Leveraging the relationship of our task to detecting contextual attributes such as negation, the conTEXT \cite{chapman2007context}  algorithm embedded in the v3NLP-Framework was augmented with a few additional entries including ``able'' and ``independent'' as asserted evidence and ``unable'' as negative evidence.

The conTEXT algorithm relies on a lexicon of evidence and accompanying clues to indicate when evidence found to the right or left of a relevant entity within a bounded window should be applied.  
We used the sentence containing an Action mention as the bounds of its context window. 
An \textit{Action Polarity} UIMA annotator was built to assign Polarity, given an Action annotation.  This annotator is downstream from the conTEXT annotator that assigned negation, assertion, conditional, hypothetical, historical, and subject attributes to named entities.   
Within conTEXT-processed entities, we assigned \textit{Unable} polarities to actions that had previously been attributed with negative and assigned \textit{Able} polarities that had previously been assigned only asserted attributes. Actions that were tagged as conditional or hypothetical were not assigned a Polarity. 

The v3NLP-Framework pipeline includes document decomposition annotators to identify sections, section names, sentences, slots and values, questions and their answers, and to a lesser extent checkboxes \cite{divita2014recognizing}. Action mentions in clinical text occur within the boundaries of each of these elements.  
ConTEXT addresses action mentions within prose, but is not relevant for action mentions found in the semi-structured constructs.  The Action Polarity annotator was thus augmented with additional rules to aid in polarity assignment based on where the mention was found. The most relevant rules are as follows:  

\begin{itemize}
    \item Action mentions that are in the slot part of a slot:value construct get their polarity assignment from positive or negative evidence in the value part of the construct. Table~\ref{tab:slotValueActionPolarityTable} provides guidelines to assigning polarity from slot:value and question and answer constructs.

    \item Action mentions that are within Goals or Education sections do not get a polarity.  The section name is known for each named entity.  For the time being, section names with ``plan,'' ``goals,'' ``education,'' ``intervention'' and ``recommendations'' qualify. These are considered to be hypothetical constructs.  The exception to this is if a goal is noted to have been met, it gets an \textit{Able} Polarity.
    

    \item Action mentions within only the value part of the slot:value construct were handled the same way as Action mentions within prose.  
    \end{itemize}
 
\subsection{Machine learning models}
\label{subsec:baseline}

We evaluated the following common machine learning-based classification methods
for our Polarity labeling task:\footnote{
    We used the implementations of each method in Scikit-Learn \cite{scikit-learn}.
}
\begin{itemize}
    \setlength{\itemsep}{1pt}
    \item Random forest (RF), using 100 estimators;
    \item Na\"{i}ve Bayes (NB), using Gaussian estimators;
    \item $k$-nearest neighbors (kNN), using $k$=5 with Euclidean distance;
    \item Support vector machine (SVM), with linear kernel;
    \item Deep neural network (DNN), using a 100-dimensional hidden layer
          followed by a 10-dimensional hidden layer.\footnote{
              We experimented with $d\in{10,100}$, and number of layers $\in {1,2,3}$.
          }
\end{itemize}

For a given Action mention $a$ contained in a Mobility description $m$, we
explored using both bag of binary unigram features\footnote{
    Binary unigram features consistently matched or outperformed unigram
    counts in our experiments.
} and word embedding features as model input. For both kinds of features, we
experimented with using the context words in $m-a$ (i.e., all words in $m$ except for the Action mention itself) only, and including the
text of the Action mention $a$. Word embedding features were calculated by averaging the embeddings of all words used (either context alone or averaging context words and Action mention words together); we used pretrained FastText
\cite{Bojanowski2017} embeddings from Wikipedia and newswire, including subword
information.\footnote{
    \url{https://fasttext.cc/docs/en/english-vectors.html}
} Where both unigram and embedding features are used, they are concatenated as a single feature vector.

\subsubsection{Feature selection}
\label{subsubsec:feature-selection}

In order to identify the best combination of features for the task, we
performed five-fold cross validation experiments on the training data.
As shown in Table~\ref{tbl:feature-selection}, we found that three model
configurations achieved statistically equivalent macro F1 in cross validation
($p\geq0.001$ with bootstrap permutation test, $R=10000$ \cite{Berg-Kirkpatrick2012}).\footnote{
    We use significance threshold $p = 0.001$ throughout this paper, as a
    conservative Bonferroni correction for multiple testing. To have sufficient
    resolution to those low threshold, we use 10,000 replicates in bootstrapping. 
}
These are RF with unigram features (78.5\% F1), the 2-layer DNN with unigram and
embedding features from context only (80.9\%), and SVM with all features, i.e.\
unigrams and embeddings with both the mobility description and Action mention
texts (81.7\%).\footnote{
    Complete results tables will be made available online.
}

\begin{table}[t]
    \centering
    {\small
    \begin{tabular}{r|ccccc}
        Features   &   NB&    RF & kNN&    SVM &  DNN\\
        \hline
        Unigrams   &41.3&\B{77.3}&\B{67.0}&   78.6 &79.8\\
        +Action    &42.1&   73.7 &56.8&   80.9 &78.0\\
        +Embeddings&41.6&   64.3 &66.3&   78.8 &\B{80.9}\\
        +Both      &\B{43.0}&   65.1 &65.2&\B{81.7}&79.6\\
    \end{tabular}
    }
    \caption{Macro F1 over Polarity classes in 5-fold cross validation
             feature selection experiments. All experiments start with binary
             unigram features using context words alone, and add Action words, embedding features from context words, or
             both (i.e., unigrams and embedding features from context and Action words combined). The best performing model configurations are marked in bold.}
    \label{tbl:feature-selection}
\end{table}

Given the class imbalance in our dataset, we also analyzed per-class performance
of each model. Interestingly, as Table~\ref{tbl:ml-xval-results} illustrates, we
found that all models except Na\"{i}ve Bayes were surprisingly robust to this
imbalance, with both SVM and DNN achieving over 76\% F1 on the smallest class
(\textit{Unable}). Across all four classes, the SVM and the 2-layer
DNN yield statistically equivalent performance ($p\geq0.001$); we therefore
use absolute macro F1 to choose SVM as the best baseline model for
comparing across approaches.

\begin{table}[t]
    \centering
    {\small
    \setlength{\tabcolsep}{2.6pt}
    \begin{tabular}{r|cccc|c}
        Model         &   Able & Unable &Unclear &None &  Macro \\
        \hline
        NB (All)      &   68.2 &   15.1 &   25.6 &   62.9 &   43.0 \\
        RF (Uni)      &   84.5 &   68.1 &   69.9 &   86.7 &   77.3 \\
        KNN (Uni)     &   73.5 &   53.3 &   62.6 &   78.5 &   67.0 \\
        SVM (All)     &\B{86.3}&   76.2 &\B{76.4}&\B{87.8}&\B{81.7}\\
        DNN (Uni+Emb) &   85.0 &\B{76.8}&   74.3 &   87.5 &   80.9 \\
    \end{tabular}
    }
    \caption{Five-fold cross validation results (F1) by class with best configurations
             of learned baselines. \textit{All} indicates using unigrams,
             embeddings, and Action mention features; \textit{Uni} indicates
             using unigram features from context words only, and \textit{Uni+Emb}
             indicates both unigram and embedding features from context words.
             The best result in each column is marked in bold.}
    \label{tbl:ml-xval-results}
\end{table}

\subsubsection{CNN model}
\label{subsec:cnn}
\begin{figure}[H]
    \centering
    \includegraphics[width=.5\textwidth]{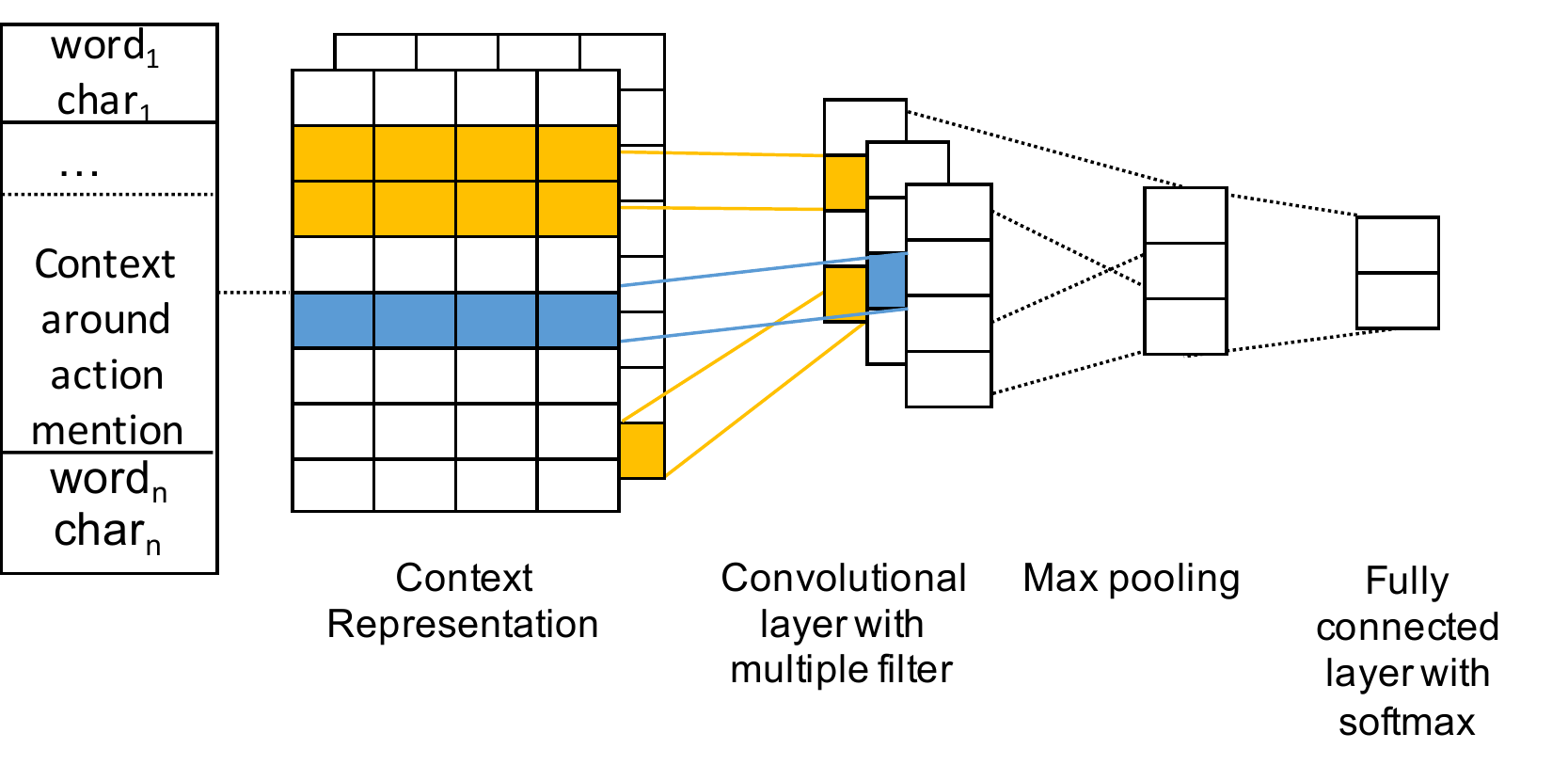}
    \caption{CNN architecture for Polarity classification.}
    \label{fig:cnn}
\end{figure}

We adopt the Convolutional Neural Network (CNN) architecture introduced in~\citet{kim2014convolutional}. In our architecture, shown in Figure~\ref{fig:cnn}, we combine word embeddings with character embeddings, to reduce the impact of out-of-vocabulary rate as opposed to using words alone. Additionally, character-level CNNs have been shown to improve the results of text classification~\cite{zhang2015character}, but the improvement is more evident with  larger data sizes.

Although our task is close to negation detection, it differs in that we do not need to detect the span of the Action: we take as inputs the Action mention and its parent mobility mention (a self-contained text span that can be considered a sentence). Unlike sequence tagging problems, where Long-Short Term Memory (LSTM) architectures would be a good fit~\cite{fancellu2016neural}, we treat the problem as a text classification task.


We experiment with character and word embeddings of the following inputs:

\begin{itemize}
    \item previous context (\textit{prev}): the set of words preceding and including the action mention.
    \item \textit{next} context: the set of words following and including the action mention.
    \item \textit{full} context: the union of $prev$ and $next$.
\end{itemize}

We also compare the impact of using character (\textit{full\_char}) or word (\textit{full\_word}) embeddings only as opposed to combining both (\textit{*\_all}), as shown in Table~\ref{tbl:cnn}. We note that relying on part of the context significantly drops the \textit{Unable} performance. However, as expected, \textit{prev} outperforms \textit{next}, given that the words preceding the Action mention carry most of the ability-related information. For the rare \textit{Unable} class, character embeddings outperform word embeddings, with F1~72.9\% on the test set; the highest across all systems.

Hyperparameters were optimized on a dev set (we used a 90/10 train/dev split), yielding a learning rate of~$0.0001$, dropout of~$0.5$, embeddings size~$100$, and Adam optimization~\cite{kingma2014adam} with L2 regularization. 

\begin{table}[t]
\centering
\small
\begin{tabular}{r|cccc|c}
Embeddings      & Able          & Unable        & Unclear       & None       & Macro         \\ \hline
prev\_all           & 82.3          & 48.7          & 31.8          & 86.9          & 62.4          \\
next\_all           & 79.3          & 32.3          & 53.5          & 82.7          & 64.9          \\
full\_all          & \textbf{87.6} & \textbf{63.4} & 65.0          & \textbf{89.4} & \textbf{76.4}   \\
full\_char     & 66.0          & 45.7          & \textbf{72.9} & 78.7          & 65.8          \\
full\_word     & 86.1          & 42.4          & 60.3          & 88.0          & 69.2
\end{tabular}
\caption{CNN performance using different inputs.}
\label{tbl:cnn}
\end{table}
\begin{table*}[t]
    \centering
    \setlength{\tabcolsep}{5pt}
    {\small
    \begin{tabular}{r|ccc|ccc|ccc|ccc|c}
        \multirow{2}{*}{System}
            &\multicolumn{3}{c|}{Able}
            &\multicolumn{3}{c|}{Unable}
            &\multicolumn{3}{c|}{Unclear}
            &\multicolumn{3}{c|}{None}
            &\multirow{2}{*}{Macro F1}\\
        &Pr&Rec&F1&Pr&Rec&F1&Pr&Rec&F1&Pr&Rec&F1&\\
        \hline
        Rule-based       &   58.3 &   71.3 &   64.2 &   20.3 &   52.2 &   29.3 &    8.8 &   12.5 &   10.3 &   80.2 &   54.2 &   64.7 &   42.1 \\
        SVM              &   83.4 &   86.8 &   85.1 &   62.1 &\B{78.3}&\B{69.2}&   63.0 &   70.8 &   66.7 &   90.0 &   84.3 &   87.0 &   77.0 \\
        CNN              &   86.0 &   89.2 &\B{87.6}&\B{72.2}&   56.5 &   63.4 &\B{81.2}&   54.2 &   65.0 &   89.0 &\B{89.7}&   89.4 &   76.4 \\
        \hline
        All (DNN chooser)&\B{87.5}&   86.3 &   86.9 &   56.7 &   73.9 &   64.2 &   66.7 &   70.8 &\B{68.7}&   90.3 &   89.5 &\B{89.9}&   77.4 \\
        SVM+CNN (Voting) &   82.3 &\B{90.8}&   86.4 &   62.1 &\B{78.3}&\B{69.2}&   62.1 &\B{75.0}&   67.9 &\B{94.5}&   82.2 &   87.9 &\B{77.9}\\
    \end{tabular}
    }
    \caption{Precision (Pr), Recall (Rec), and F1 for each model evaluated on
             the test set. Top rows are individual models, bottom rows are ensembled
             results. The best result in each column is marked in bold.}
    \label{tbl:test-results}
\end{table*}

\subsection{Ensemble models}
\label{subsec:ensemble}

Ensembling methods have been shown to improve performance in a variety of
classification tasks \cite{Buda2018}, including in class-imbalanced tasks
\cite{Ju2018}. In order to combine the strengths of each modeling approach,
we therefore experimented with ensembling all three systems, using two
ensembling strategies:

\textbf{Majority voting}
Predictions from the single best configurations of the SVM and CNN models\footnote{
    Adding rule-based predictions degraded performance in this case.
} were combined to make a single decision. When the systems agreed, that label was chosen as output; in the case of disagreement, we chose the predicted
label that was \textit{less} frequent in training data, in order to
prefer the strengths of individual models on rare classes.

\textbf{DNN chooser}
Predictions from all three systems (rule-based and the best pretrained SVM and CNN models)\footnote{
    For the chooser, adding rule-based predictions consistently improved results over just SVM and CNN.
} were passed as inputs to a DNN with
a single 10-unit hidden layer.\footnote{
    Experiments with a 64-unit hidden layer, to cover all possible label
    combinations, yielded the same results in cross validation.
} In order to compensate for the class imbalance in our dataset, which would lead to preferring the CNN due to its higher precision, we identified all training samples that the three models disagreed on and grouped them by label, and identified the smallest of these disagreement sets. We then sampled no more than twice this number of points from each disagreement set, yielding a training sample of 182 points.

Using this downsampled training set, we trained the DNN to predict
which, if any, of the systems chose the correct answer. As multiple systems
may have made the correct prediction, this is a multi-label classification
task. At test time, the system with highest probability output from the
DNN was chosen as the reference decision for the final classification.

We also experimented with three approaches to predict the final class directly:
using a DNN with the predictions of each system as input, using an SVM with
predictions as input, and adding rule-based and CNN predictions as additional
features to the SVM with lexical features. All variants underperformed the
chooser in cross validation experiments on training data, thus we omit them
from our results.

\section{Results}
\label{sec:results}

\begin{figure*}[t]
\centering
\begin{subfigure}[b]{.45\textwidth}
    \centering
    \includegraphics[width=\linewidth]{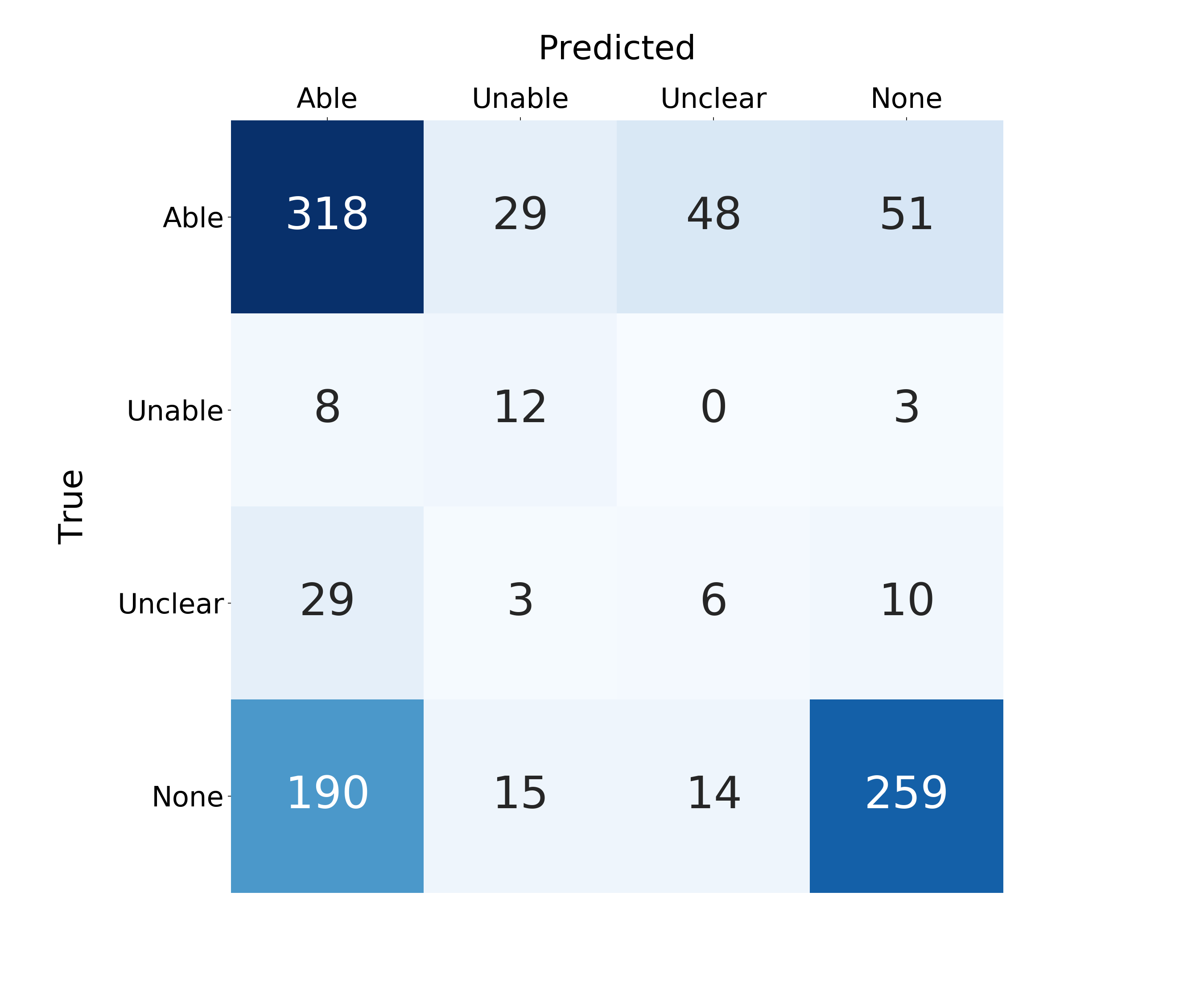}
    \caption{Rule-based}
\end{subfigure}%
\begin{subfigure}[b]{.45\textwidth}
    \centering
    \includegraphics[width=\linewidth]{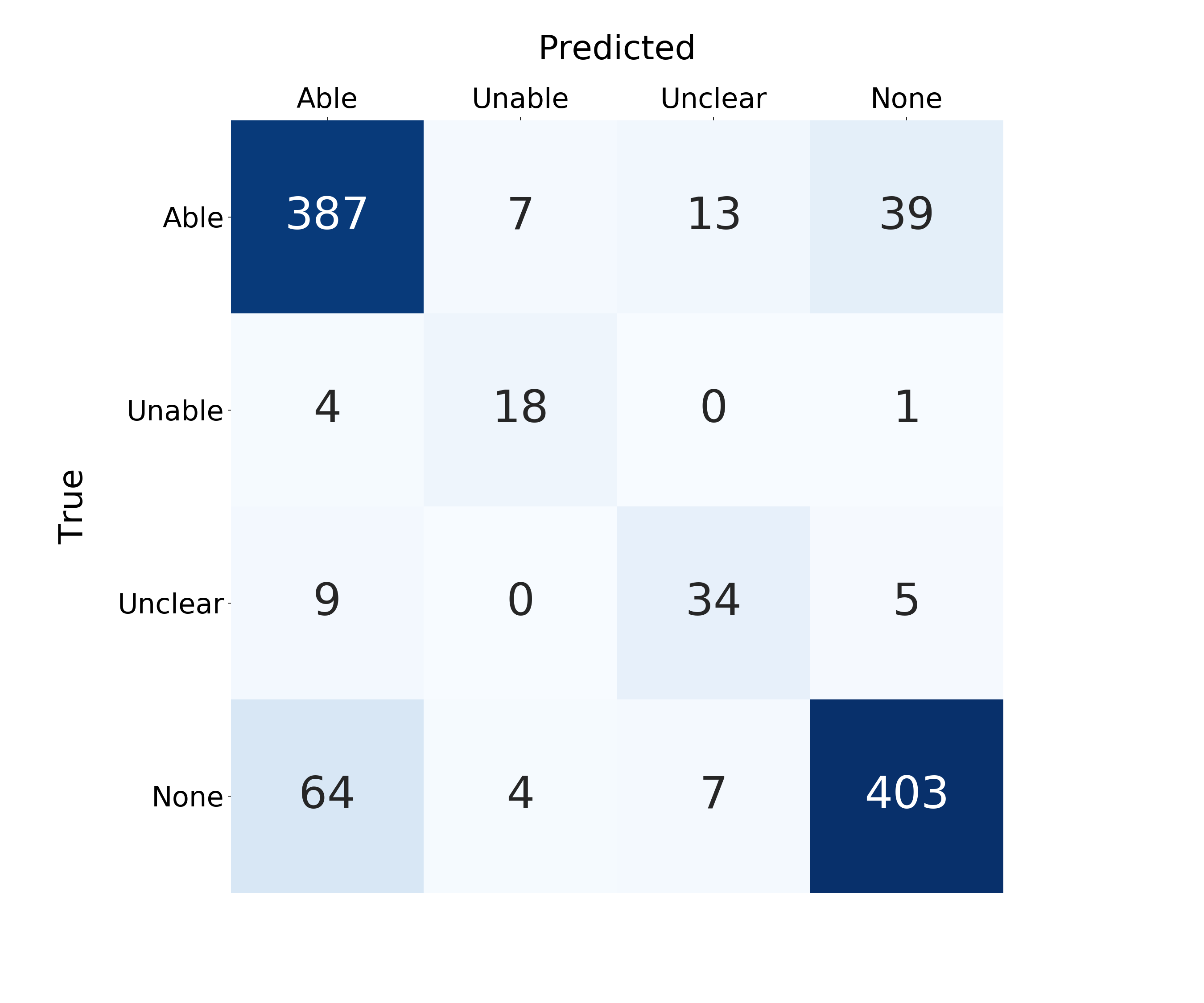}
    \caption{SVM}
\end{subfigure}
\begin{subfigure}[b]{.45\textwidth}
    \centering
    \includegraphics[width=\linewidth]{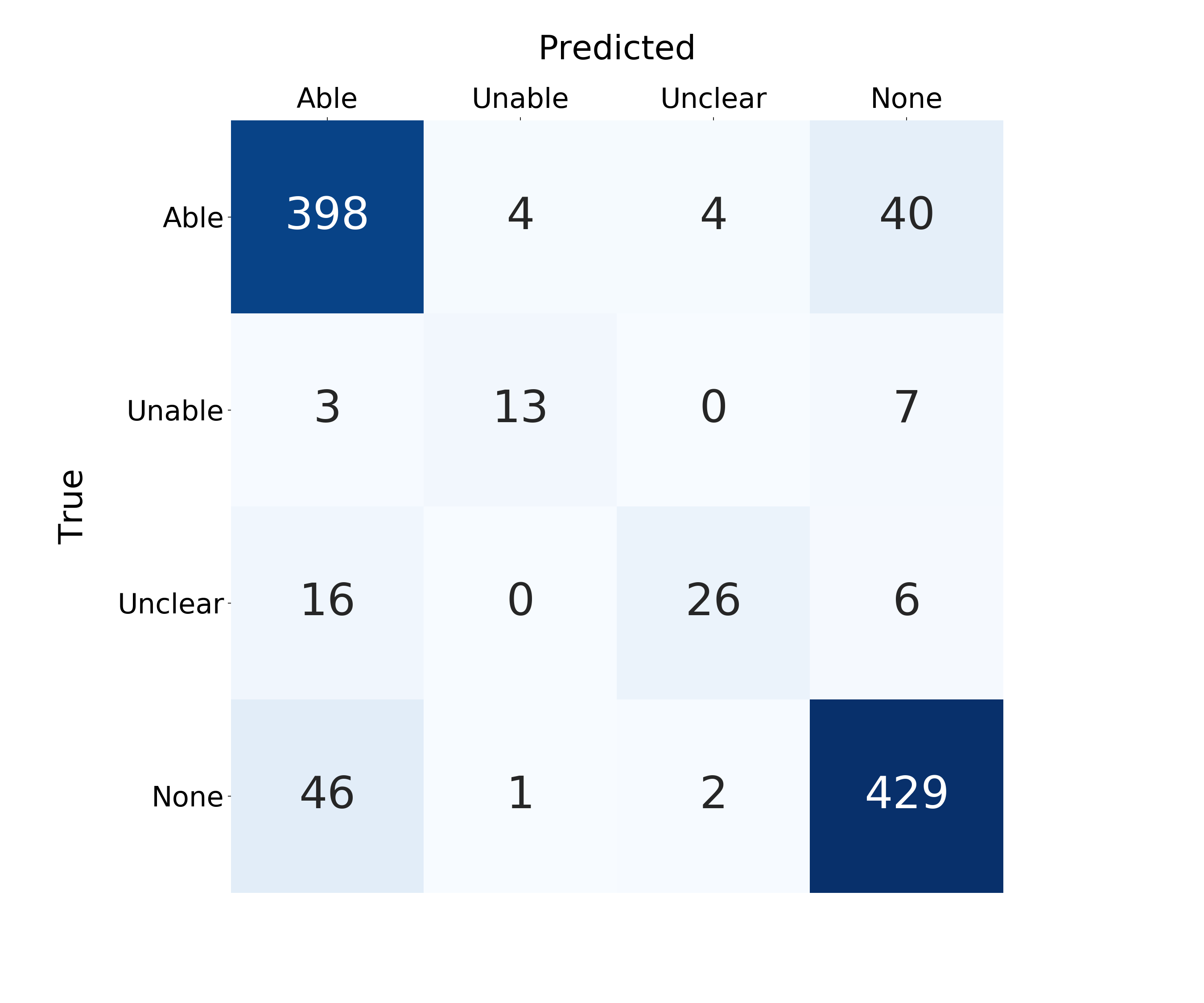}
    \caption{CNN}
\end{subfigure}%
\begin{subfigure}[b]{.45\textwidth}
    \centering
    \includegraphics[width=\linewidth]{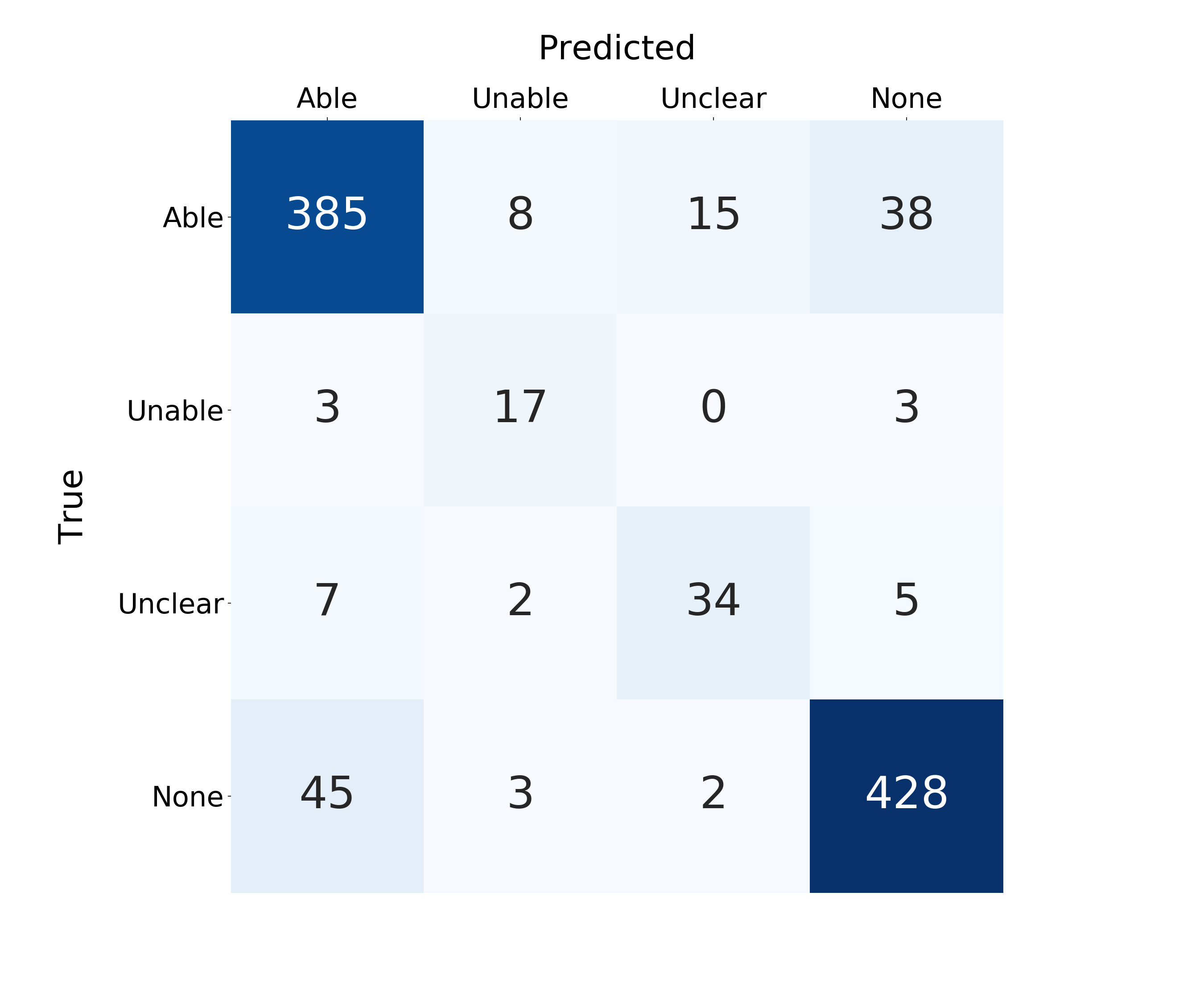}
    \caption{Ensemble (DNN chooser)}
\end{subfigure}
\caption[short]{Confusion matrices for results on the test set.}
\label{fig:confusion-matrices}
\end{figure*}

The test results of the systems we compared are given in Table~\ref{tbl:test-results}. The ensembled systems achieve the best overall performance, with 77.4\% macro F1 with the DNN chooser and 77.9\% with majority voting. Due in large part to the class imbalance in the dataset, the SVM, CNN, and ensemble methods do not yield statistically significantly different results in most cases ($p>0.001$), although the voting ensemble does produce significantly higher precision on \textit{None} samples than other methods ($p\ll0.001$).

While performance is considerably better on the more frequent \textit{Able} and \textit{None} classes, the learned systems achieve good results on \textit{Unclear} and the very rare \textit{Unable}. Figure~\ref{fig:confusion-matrices} shows the confusion matrices for all systems. The most common confusions are with \textit{Able} and \textit{None}, with only a small number of false positives for \textit{Unable} and \textit{Unclear} and no confusion between the two in the machine learning approaches.

Comparing between individual systems, the CNN is best at making the important distinction between \textit{Able} and \textit{Unable}. It consistently achieves high precision across all classes, but suffers large drops in recall for the rare labels. The SVM model reverses this tradeoff, yielding high recall for \textit{Unable} and \textit{Unclear}, but much lower precision. The ensembled methods are able to strike a good middle ground, keeping the high recall of the SVM without sacrificing too much of the CNN's precision.

\section{Discussion}
\label{sec:discussion}
As is evident from the results, correctly classifying the minority classes \textit{Unable} and \textit{Unclear} is not trivial. This is not only caused by the lack of data for training those classes, but in the case of \textit{Unclear}, also by its semantic ambiguity -- even for humans.

An important area of confusion is when actions are hypothetical, as is the case for plans, recommendations or feelings towards an action (e.g. \texttt{eager to walk}), which should all be tagged as \textit{None}. Semantic problems can also arise around the use of an assistive device. In the following synthetic example, the annotated polarity is \textit{Able}: \texttt{she is unable to ambulate more than a few feet without support}. Without the mention of assistance, it would have been \textit{Unable}. In future work, assistance mentions will be modeled explicitly to better capture this.

Overall, we obtain models that perform well across the board, where each approach has different strengths as illustrated in Figure~\ref{fig:confusion-matrices}. Out of the 955 test instances, the rule-based approach classifies 37 correctly that no other system got right. Likewise, SVM and CNN have 27 and 25 unique true positives, respectively. 46 instances get misclassified by all classifiers. The ensemble is able to pick up on 31 of the unique true positives from the machine learning systems, but consistently ignores valid suggestions from the rule-based approach. This suggests that different ensembling parameters should be considered to take better advantage of the rule-based system's strengths.

Below, we discuss system-specific observations in more detail.

\subsection{Rule-based}
The following failures were observed in the training and testing output:

\textbf{Scoping negation} 
The scope for assigning negation attribution was set to be within sentential boundaries.  Ideally, the scope should be tighter at the major phrase level.  However, v3NLP-Framework does not currently employ a dependency graph parser. 
Breaking on phrasal boundaries was not successful, primarily due to the inability to distinguish between list markers such as commas, coordinating conjunctions (and/or), and true scope limiting phrasal boundaries.  
Several false negatives were due to the incorrect \textit{Unable} assignment because of negation scoping.

\textbf{Identifying variants of slots and values accurately}
Negation and assertion assignment are dependent upon whether the action is within prose, a slot or a value.  A number of errors were due to multiple slot:value constructs within the same line making it difficult identifying the values, and/or nested constructs  (i.e.,\ the value of a slot:value construct was also a slot:value construct).

\textbf{Nested sections}
A number of missed \textit{None} errors were the result of mis-identifying what section the annotation was within, and picking up an inner section name.
Several other issues arose from the use of spaces as delimiters between slots and values, as well as slots and values embedded within bulleted lists. 

\textbf{Pertinent negatives} \cite{divita2014recognizing}
A statement where the action mention had clear negative evidence really meant the patient could perform an action.  For example, \clinquote{no trouble walking}. An easy amelioration would be to gather constructs like ``no trouble'' and add them to the assertion evidence lexicon.  

\subsection{Machine learning}
The machine learning systems are prone to failures in sentences that have multiple Action mentions, if their Polarity differs. This is because the systems do not take into account sentence structure. Similarly, sentence length seems to have a negative effect on performance, as it dilutes the information salient to the focus mention. In future work, we would limit the context information to exclude other mentions' contexts, add parse tree information relevant to the focus mention, or improve the neural network architecture to better model the sequential nature of the data. 

The models would also benefit from better capturing semantic similarity. An example would be \texttt{Pt.~is fearful to start walking again} (class: \textit{None}), where the modality expressed by \textit{fearful} might not have been learned from the training data. Additionally, lemmatization, stemming and character embeddings can blunt the impact of such unseen tokens, but using embeddings from large corpora would be more robust. 

Finally, one potential limitation in our machine learning results is our use of pretrained embeddings from web text. As \citet{newman2018embedding} show, when only a small amount of text from the target domain is available, out-of-domain embeddings can roughly match performance with in-domain embedding features; however, developing or tuning more targeted word embeddings for use in this dataset is a useful area of future work.

\subsection{Generalizability}
It is important to note that the dataset used in this study was
derived from one specialty -- Physical Therapy -- within a single
institution -- the NIH Clinical Center. Thus, the texts analyzed are likely to
be more homogeneous than would be a broader dataset. Evaluating
generalization of our findings to free text from other healthcare subdomains
and other institutions, and describing ways in which performance assertions
vary between these sources, is a valuable area of future work.

\section{Conclusion}
\label{sec:conclusion}

We have presented an evaluation of several approaches for the task of classifying whether a given description of an individual performing an activity indicates that they are able to perform it, unable, unclear, or insufficient information to determine. We found that machine learning approaches with lexical features perform surprisingly well on the task, including detecting the rarer labels of \textit{Unable} and \textit{Unclear}, and that an ensembled approach sets a strong baseline of 77.9\% macro F1 for our dataset. In-depth analysis of system errors suggested several intriguing problems for future work. For instance, we intend to investigate hybrid models and test how information related to report formatting, section structure, slot info and assistive devices could improve the performance. To clarify the confusion of a patient's ability, we need models that can differentiate between factual and hypothetical statements (e.g. \texttt{Pt can run} vs. \texttt{Pt dislikes running}). Additionally, we would like to incorporate contextual representations such as ELMo~\cite{elmo} and BERT~\cite{bert} into our models.

To our knowledge, this is the first work expanding on the problem of clinical negation detection to complex interactions between individuals and their environments. This work joins a growing body of research on application of NLP techniques to information about activity performance and role participation, and identifies several research challenges in adapting NLP methods to this new domain.

\section*{Acknowledgments}

The authors would like to thank Pei-Shu Ho, Jonathan Camacho Maldonado,
and Maryanne Sacco for discussions about error analysis, and our anonymous reviewers
for their helpful comments. This research was supported in part by the Intramural Research Program of the
National Institutes of Health, Clinical Research Center and through an
Inter-Agency Agreement with the US Social Security Administration.

\bibliography{ref}
\bibliographystyle{acl_natbib}

\end{document}